\documentclass{article}
\usepackage{ijcai24}

\usepackage{natbib}

\usepackage{times}
\usepackage{soul}
\usepackage{url}
\usepackage[hidelinks]{hyperref}
\usepackage[utf8]{inputenc}
\usepackage[small]{caption}
\usepackage{graphicx}
\usepackage{amsmath}
\usepackage{amsthm}
\usepackage{booktabs}
\usepackage{algorithm}
\usepackage[switch]{lineno}

\usepackage{CJKutf8}

\usepackage{amssymb}
\usepackage{amsthm}
\usepackage{enumitem}
\usepackage{picinpar}
\usepackage[noend]{algpseudocode}
\usepackage{subfigure}
\usepackage{multirow}
\usepackage{color}
\usepackage{balance}
\usepackage{tabularx}
\usepackage{enumitem}
\usepackage{hhline}
\usepackage[normalem]{ulem}
\usepackage{wrapfig}
\usepackage{bm}
\usepackage{adjustbox}

\usepackage{booktabs} 
\usepackage{makecell} 
\usepackage{multirow} 
\usepackage{multicol} 
\usepackage{graphicx} 
\usepackage{bbding}

\newcommand{\bN}{\ensuremath{\mathcal{N}}}

\renewcommand{\vec}[1]{\ensuremath{\mathbf{#1}}}
\newcommand{\stitle}[1]{\vspace{2mm} \noindent {\bf #1}}
\newcommand{\eg}{{\it e.g.}}

\newcommand{\ie}{{\it i.e.}}
\newcommand{\etc}{{\it etc.}}
\newcommand{\wrt}{w.r.t. }

\newcommand{\method}{}

\newcommand{\eat}[1]{}

\title{Advancing Graph Representation Learning with Large Language Models: \\
A Comprehensive Survey of Techniques}

\author{
Qiheng Mao$^1$
\and
Zemin Liu$^2$\thanks{Corresponding author}\and
Chenghao Liu$^3$\footnotemark[1]\and
Zhuo Li$^4$\And
Jianling Sun$^1$\\
\affiliations
$^1$Zhejiang University; \hspace{1em}
$^2$National University of Singapore; \\
$^3$Salesforce Research Asia; \hspace{1em}
$^4$State Street Technology (Zhejiang) Ltd.
\emails
\{maoqiheng, lizhuo, sunjl\}@zju.edu.cn,
zeminliu@nus.edu.sg,
chenghao.liu@salesforce.com
}

\begin{document}

\maketitle

\begin{abstract}
The integration of Large Language Models (LLMs) with Graph Representation Learning (GRL) marks a significant evolution in analyzing complex data structures. This collaboration harnesses the sophisticated linguistic capabilities of LLMs to improve the contextual understanding and adaptability of graph models, thereby broadening the scope and potential of GRL. Despite a growing body of research dedicated to integrating LLMs into the graph domain, a comprehensive review that deeply analyzes the core components and operations within these models is notably lacking. Our survey fills this gap by proposing a novel taxonomy that breaks down these models into primary components and operation techniques from a novel technical perspective. We further dissect recent literature into two primary components including knowledge extractors and organizers, and two operation techniques including integration and training stratigies, shedding light on effective model design and training strategies. Additionally, we identify and explore potential future research avenues in this nascent yet underexplored field, proposing paths for continued progress.
\end{abstract}

\section{Introduction}

The prevalence of graph data has spurred the need for advanced network analysis. A crucial component of this analysis is Graph Representation Learning (GRL). This field primarily focuses on graph embedding \citep{cai2018comprehensive,perozzi2014deepwalk,grover2016node2vec}, Graph Neural Networks (GNNs) \citep{wu2020comprehensive,kipf2017semisupervised,velickovic2018graph}, and graph Transformers \citep{min2022transformer,rong2020self}, which aim to represent graph elements (\eg, nodes) into low-dimensional vectors. 
The advances in GRL have profound implications across various domains, reflecting its growing importance in extracting meaningful insights from complex data.

With the advancements in machine learning techniques, the rapidly evolving field of artificial intelligence is opening new avenues for research, particularly with the rise of Large Language Models (LLMs) \citep{zhao2023survey}. Known for their impressive performance in Natural Language Processing (NLP), LLMs show potential for wide-ranging applications in various areas beyond NLP \citep{li2023blip,luo2022biogpt}. Their skill in identifying complex patterns in large datasets leads to an important question: \emph{Can the powerful capabilities of LLMs be harnessed to significantly enhance graph representation learning?}

\stitle{Advancing GRL with LLMs.}
The flexibility of LLMs, shown in their advanced ability to understand and generate human language, makes them a key tool for analyzing complex data structures, including graphs. By combining the advanced analysis power of LLMs with graph data, we have a promising chance to bring the text knowledge and wide-ranging application skills of LLMs into graph representation learning. This combination not only gives graph models a better grasp of context and meaning but also improves their ability to adapt to different situations. This expansion increases the potential and usefulness of GRL in understanding the complexity and connectedness of data in various fields \citep{chen2023exploring,tang2023graphgpt,he2023explanations}.

As a result, the impressive capabilities of Large Language Models (LLMs) have led to a growing number of research efforts focused on integrating LLMs into the graph domain \citep{he2023explanations,chen2023exploring,chen2023label,ye2023natural}. This research is advancing a variety of applications in graph-related tasks, including node classification \citep{chen2023exploring,huang2023can}, graph reasoning \citep{wang2023can,guo2023gpt4graph}, molecular graph analysis \citep{su2022molecular,liu2022multi}, \etc\ 
Despite extensive exploration in this area, there is a noticeable lack of systematic reviews summarizing the progress of graph representation learning with LLMs, particularly from a technical standpoint of \emph{how to design a graph foundation learning model with the assistance of LLMs}. Addressing this gap is crucial as it can deepen our understanding of the current developments in graph learning models enhanced by LLMs and aid in creating more effective future graph foundation models.

\begin{figure}[t]
\centering
\includegraphics[width=0.9\linewidth]{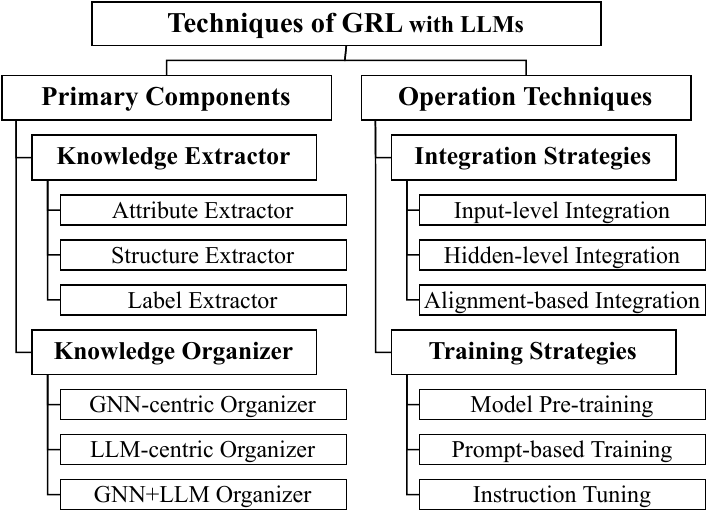}
\caption{The techniques of GRL with LLMs.}
\label{fig:taxonomy}
\vspace{-3mm}
\end{figure}

\stitle{Our survey.}
To bridge this gap and enhance understanding of the recent literature on GRL with LLMs from a technical perspective, this paper systematically reviews and summarizes the technical advancements in GRL with LLMs. Our goal is to enable readers to thoroughly analyze this technique and understand how to design GRL models assisted by LLMs. This, in turn, will contribute to advancing the field of graph representation learning.

To structure this survey, we begin by decomposing the techniques of the existing literature on GRL with LLMs into two \emph{primary components}: \emph{knowledge extractors} and \emph{knowledge organizers}, based on their respective roles within the overall model. Specifically, with a focus on graph representation learning, knowledge extractors (Section \ref{sec.knowledge-extractor}) are concerned with extracting knowledge from graph data (\eg, graph encoders), while knowledge organizers (Section \ref{sec.knowledge-organizer}) are responsible for arranging and storing this knowledge (\eg, multi-layer transformers).

In addition to these major components in a GRL model enhanced with LLMs, we also provide a comprehensive overview of the existing \emph{operation techniques} used for model management. This includes \emph{integration strategies} for combining graph learning models (\eg, GNNs) with LLMs and \emph{training strategies} for effectively training the unified model.
With knowledge extractors and organizers as the foundation, the operation techniques primarily address how to manage and integrate modules by integration strategies (Section \ref{sec.integration-strategies}) to ensure their seamless combination, and how to achieve effective and efficient training by training strategies (Section \ref{sec.training-strategies}).

The overall taxonomy is depicted in Figure~\ref{fig:taxonomy}, which is divided into two main branches to highlight the primary components and operation techniques. 
Building on this framework, we delve into each branch in detail, summarizing the existing literature to showcase the advancements in both major components and operation techniques. Specifically, we compare the related literature from the perspective of their configurations in constructing their entire model, providing readers with a deeper understanding of the various structures, and further illustrate the details in Table \ref{tab:modelsummary}. Additionally, we highlight potential future research directions that merit exploration to further advance GRL with LLMs.

\stitle{Relationships with existing surveys.}
The recent focus on LLMs applied to graph data has prompted several research surveys to review the current literature \citep{jin2023large,liu2023towards,li2023survey, zhang2023large}. 
While sharing similar investigation directions, these surveys lack a focus on general graph representation learning domain.
They predominantly classify the literature based on LLMs' roles (\eg, as predictor, aligner, \etc) and specific application scenarios, overlooking a holistic overview and detailed analysis of graph foundation models' structural framework.
Our work sets itself apart by treating GRL with LLMs as an integrated model and meticulously breaking it down from a technical standpoint into several essential components and operations. To achieve this, we introduce a novel taxonomy to structure the literature and delineate the advancements from each perspective. This approach offers a comprehensive and detailed analysis of these models, equipping readers with the knowledge to design their own graph foundation models. 

\stitle{Contributions.} To summarize, our contributions are three-fold.
(1) We provide a comprehensive analysis of research efforts that integrate LLMs into GRL, viewing them as unified models. This includes a detailed breakdown of these models into essential components and operations from a novel technical perspective.
(2) We propose a novel taxonomy that categorizes existing models into two main components: knowledge extractors and organizers, according to their functions within the model. This taxonomy also covers operation techniques, including integration and training strategies, to offer readers a deeper understanding of graph foundation models.
(3) Utilizing our taxonomy, we identify and discuss potential future research directions in this emerging yet underexplored field, suggesting avenues for further development.

\stitle{Organization.} 
The structure of this survey is outlined as follows. Section \ref{sec.background} provides an introduction to the background relevant to this survey. 
Sections \ref{sec.knowledge-extractor} and \ref{sec.knowledge-organizer} detail the \emph{primary components} identified in the existing literature, focusing on knowledge extractors and knowledge organizers, respectively. 
Sections \ref{sec.integration-strategies} and \ref{sec.training-strategies} discuss the \emph{operation techniques} highlighted in the literature, specifically addressing integration strategies and training strategies. 
Section \ref{sec.future-directions} explores potential future research directions in line with our organizational framework. The survey concludes with Section \ref{sec.conclusions}.

\begin{table*}[!htbp]
    \centering
    \renewcommand{\arraystretch}{1.5} 
    \resizebox{\textwidth}{!}{%
    \begin{tabular}{@{}l|ccc|c|c|c|c@{}}
        \toprule
        \multirow{2}*{\textbf{Approaches}} &
   \multicolumn{3}{c|}{\textbf{Knowledge Extractor}}  & \multirow{2}*{\makecell*[c]{\textbf{Knowledge} \\ \textbf{Organizer}}}  & \multirow{2}*{\makecell*[c]{\textbf{Integration} \\ \textbf{Strategies}}}  & \multirow{2}*{\textbf{Training Strategies}} & \multirow{2}*{\textbf{Tasks}} \\ \cmidrule{2-4}
       & \textbf{Attribute Extractor} & \textbf{Structure Extractor} & \textbf{Label Extractor} &  &  & \\\midrule\midrule
        TAPE \citep{he2023explanations} & Feature, Text & original & \Checkmark & GNN & hidden-level & Prompting & Node \\ 
        Chen et al. \citep{chen2023exploring} & Feature, Text & no graph, subgraph, original & \Checkmark & GNN/LLM & input-level & Prompting & Node \\ 
        ConGraT \citep{brannon2023congrat} & Feature  & original & \XSolidBrush & GNN+LLM & alignment-based & Pretraining & Node, Link \\
        GraphGPT \citep{tang2023graphgpt} & Feature & subgraph,original & \XSolidBrush & LLM & alignment-based & Pretraining, Prompting, Instruction Tuning & Node \\ 
        G-Prompt \citep{huang2023prompt} & Feature & subgraph & \XSolidBrush & LLM & input-level & Prompt Tuning & Node \\ 
        ENG \citep{yu2023empower} & Feature & structure refinement & \Checkmark & GNN & input-level & Prompting & Node \\ 
        Sun et al. \citep{sun2023large} & Feature & structure refinement & \Checkmark & GNN & - & Prompting & Node \\ 
        GraphText \citep{zhao2023graphtext} & Feature & structure refinement & \XSolidBrush & LLM & input-level & Prompting & Node \\ 
        GLEM \citep{zhao2022learning} & Feature & original & \Checkmark & GNN+LLM & alignment-based & - & Node \\
        LLM-GNN \citep{chen2023label} & Feature & original & \Checkmark & GNN & - & - & Node \\
        GRAD \citep{mavromatis2023train} & Feature & original & \XSolidBrush & LLM & alignment-based & - & Node \\
        G2P2 \citep{wen2023prompt} & Feature & original & \XSolidBrush & GNN+LLM & alignment-based & Prompt Tuning & Node \\
        Patton \citep{jin2023patton} & Feature & original & \XSolidBrush & GNN+LLM & hidden-level & Pretraining & Node, Link \\ 
        GALM \citep{xie2023graph} & Feature & original & \XSolidBrush & GNN & - & Pretraining & Node, Link \\ 
        SimTeG \citep{duan2023simteg} & Feature & original & \XSolidBrush & GNN & - & Pretraining & Node, Link \\ 
        OFA \citep{liu2023one} & Feature & original & \XSolidBrush & GNN & input-level & Prompt Tuning & Node, Link, Graph \\ 
        InstructGLM \citep{ye2023natural} & - & subgraph & \XSolidBrush & LLM & input-level & Pretraining, Prompting, Instruction Tuning & Node \\ 
        GPT4Graph \citep{guo2023gpt4graph} & - & no graph, subgraph & \XSolidBrush & LLM & input-level & - & Node, Graph \\ 
        Hu et al. \citep{hu2023beyond} & - & no graph, subgraph & \XSolidBrush & LLM & input-level & Prompting & Node, Link, Graph \\ 
        Huang et al. \citep{huang2023can} & - & no graph & \XSolidBrush & LLM & input-level & Prompting & Node \\ 
        \bottomrule
    \end{tabular}%
    }
    \caption{The technique summarization of existing most-related literature on graph representation learning with LLMs.}
    \vspace{-3mm}
    \label{tab:modelsummary}
\end{table*}

\section{Background} \label{sec.background}

\stitle{Graph representation learning and GNNs.}
Graph representation learning seeks to transform elements of a graph into low-dimensional vector representations, ensuring the preservation of the graph's structure. Currently, the primary models being extensively investigated for advancement in this field by LLMs are GNNs\footnote{Note that, since GNNs are the main models extensively investigated for enhancement in this field with the aid of LLMs, in this survey, ``graph learning models'' specifically refer to GNNs.}.  The fundamental process in GNNs involves recursive neighborhood aggregation, a method that compiles information from neighboring nodes to refine and update the representation of a specific node. 
Formally, let $\phi_g(\cdot;\theta_g)$ denote a GNN architecture parameterized by $\theta_g$. In the $l$-th layer, the representation of node $v$, \ie, $\vec{h}^l_v\in\mathbb{R}^{d_l}$, can be calculated by
\begin{equation}\label{eq.gnn}
    \vec{h}_v^l = \textsc{Aggr}(\vec{h}_v^{l-1},\{\vec{h}_i^{l-1}: i \in \bN_v\}; \theta^l_g),
\end{equation}
where $\bN_v$ is the neighbors set of node $v$ and $\textsc{Aggr}(\cdot;\theta^l_g)$ is the aggregation function parameterized by $\theta^l_g$ in layer $l$. 

\stitle{Large language models.}
LLMs are a category of natural language processing models known for their enormous size, often comprising billions of parameters and being trained on extensive datasets \citep{zhao2023survey}. 
These models represent a significant advancement over earlier, smaller Pre-trained Language Models (PLMs) \citep{gao2021making} in both size and capabilities, covering a wide range of tasks including understanding \cite{zhu2023minigpt} and generating \citep{madani2023large} for natural language processing.

Central to LLMs is the use of the Transformer architecture \citep{vaswani2017attention}, which allows for efficient processing of large datasets. This architecture has facilitated the development of both non-autoregressive models like BERT, which focus on understanding through tasks like masked language modeling \citep{zaken2022bitfit}, and autoregressive models like GPT-3, which excel in generation tasks through next-token prediction \citep{yang2019xlnet}.

\section{Knowledge Extractor} \label{sec.knowledge-extractor}

\begin{figure}[t]
\centering
\includegraphics[width=\linewidth]{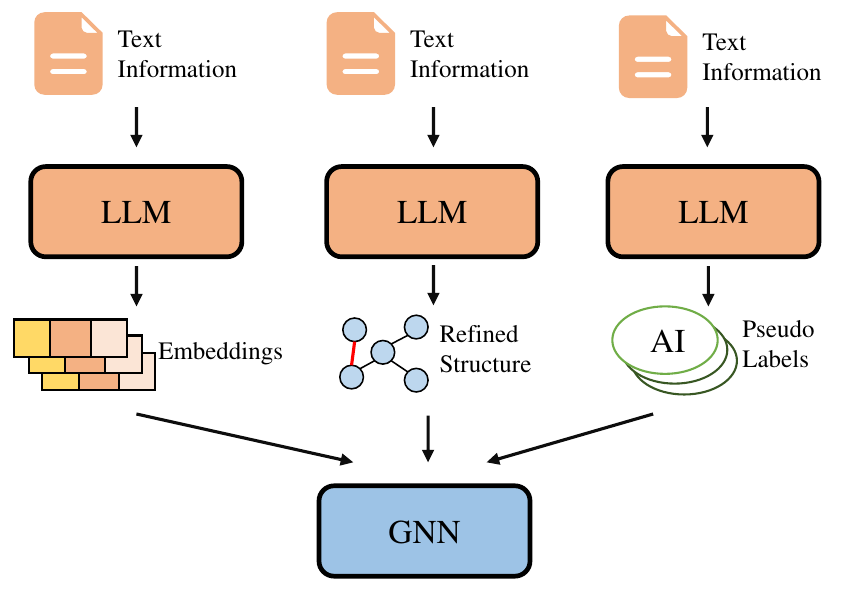}
\caption{The illustration of graph knowledge extractors on attribute, structure and label information with LLMs. 
}
\vspace{-3mm}
\label{fig:extractor}
\end{figure}

The concept of a knowledge extractor involves extracting and encoding essential information from graph data \wrt graph learning models or LLMs, ensuring that the resulting representations faithfully reflect the graph's interconnected nature. In graph representation learning, graph data typically encompasses information from three dimensions: graph attributes, graph structures, and label information. Consequently, in this section, we will explore various types of knowledge extractors, examining them through the lenses of graph attributes, structures, and labels. An illustration of the knowledge extractor can be found in Figure \ref{fig:extractor}.

\subsection{Attribute Extractor}
Attribute Extractor focus on extracting and enhancing patterns from node and edge attributes, usually leveraging LLMs to interpret and enrich the textual or numerical data associated with graph components. This involves using LLMs to generate more complete textual statements and encode more comprehensive semantic features, corresponding to the Text-level Extractor and Feature-level Extractor, respectively.

\stitle{Text-level Extractor.}
Textual-level Extractor aims to leverage the generative prowess of LLMs to augment the incomplete textual attributes. This method encompasses generating extensive descriptive or explanatory text, thus amplifying the information value of the original graph data from a textual standpoint. The resultant feature set is imbued with richer semantic depth. 
For example, \method{TAPE} \citep{he2023explanations} uses prompt to manipulate LLM to generate additional explanatory texts, with the aim of more effectively encoding text semantic information into initial node features for GNNs. 
Additionally, \cite{chen2023exploring} employs Knowledge-Enhanced Attention to prompt LLMs to generate relevant knowledge entities and textual descriptions, thereby enhancing the richness of textual information. 
These two works demonstrate that when graph data contains textual attributes, LLMs can effectively extract the textual information using their extensive knowledge repository and generative capabilities.

\stitle{Feature-level Extractor.}
The Feature-level Extractor leverages LLMs to encode the textual representations of nodes in GRL, moving beyond the traditional word embedding methods like Word2Vec \citep{mikolov2013efficient}. 
By utilizing the advanced capabilities of LLMs for more nuanced and context-aware feature encoding, the effectiveness of graph representation can be significantly enhanced. 
Due to its direct effectiveness, the feature-level extractor has been adopted by multiple studies \citep{duan2023simteg, xie2023graph}.
Using the feature-level extractor, OFA \citep{liu2023one} converts all nodes, edges and task information into human-readable texts and uniformly encodes them across various domains. It proposes a novel GRL framework capable of utilizing a singular graph model for solving diverse tasks from cross-domain datasets with the help of LLMs.

\subsection{Structure Extractor}
Knowledge extraction from graph structures plays a pivotal role in GRL. 
The structure extractor with LLMs in this arena manifests in two key ways: noisy graph structure refinement based on LLMs and graph structure utilization with LLMs. 
These approaches introduce innovative solutions for extracting graph structures, enhancing both the utilization and encoding processes of the structure knowledge.

\stitle{Graph Structure Refinement.}
Graph Structure Learning (GSL) represents an advancement over traditional GRL, which primarily focuses on concurrently optimizing both the graph structure and node representations, to enhance the overall effectiveness of the representations by improving the quality of the graph structure. In the context of Text-Attributed Graphs (TAGs), where nodes are associated with rich textual information, LLMs offer a unique opportunity to enhance the graph structure from a semantic standpoint.

Based on the observation that TAGs often contain numerous irrational or unreliable edges, \cite{sun2023large} enables the LLM to assess the semantic-level similarity between pairs of nodes in the TAGs through meticulously crafted prompts with textual attributes. 
\method{ENG} \citep{yu2023empower} leverages LLMs for generating effective nodes and corresponding graph structures in TAGs, enhancing GNN models' performance in few-shot scenarios.
Both of the aforementioned methods leverage the powerful textual semantic representation capabilities of LLMs to uncover implicit, effective graph structural relationships, thereby enhancing the efficacy of structural information in graph representation learning and boosting GSL.

\stitle{Graph Structure Utilization.}
Integrating LLMs into GRL expands the utilization of existing graph structures. This is accomplished by transforming graph structures or contexts into natural language descriptions \citep{huang2023can, guo2023gpt4graph}. Utilizing LLMs' extensive capabilities in representing language inputs allows for effective information extraction, reducing dependency on graph-structured inputs.

Although the input to LLMs does not depend on graph structures, enhancing their performance can be achieved by articulating certain graph structural information through natural language descriptions \citep{chen2023exploring}.
\method{InstructGLM} \citep{ye2023natural} creates templates to describe each node's local ego-graph structure (up to 3-hop connections), while \method{GraphGPT} \citep{tang2023graphgpt} introduces special GraphTokens, refined through Graph Instruction Tuning, to convey graph structural information, significantly reducing the length of token sequences required to describe structural information.
In addition to converting the original graph information into natural language, \method{GraphText} \citep{zhao2023graphtext} introduces a syntax tree-based approach to convert graph structure into text sequences. 
Although these techniques can furnish LLMs with partial graph-related information, the constraint on input length precludes a comprehensive representation of the full graph information. The most effective method for feeding graph information into LLMs remains a subject of ongoing investigation.

\subsection{Label Extractor}

A key obstacle in GRL, especially in real-world applications, is the scarcity of high-quality annotated data. Label information is crucial for these models to efficiently extract knowledge from graph data. The zero-shot capabilities of LLMs, derived from their extensive parameters and vast training, allow for accurate label inference using the textual features of the nodes, providing a practical solution for ongoing data labeling in GRL.

\method{LLM-GNN} \citep{chen2023label} utilizes \method{GPT-3.5} \citep{brown2020language} with a hybrid prompt for zero-shot input, annotating selected nodes and using confidence scores to discard low-quality labels. This strategy allows GNNs to perform well in node classification tasks without real labels. And the effectiveness of using LLMs for label annotation has also been validated in \citep{zhao2022learning,chen2023exploring}. 
However, current annotation methods primarily rely on LLMs' ability to understand textual features, without considering the structural relationships between nodes. If structural information could be incorporated into the LLM input, the quality of the labels is likely to be further improved.

\section{Knowledge Organizer}  \label{sec.knowledge-organizer}

\begin{figure}[t]
\centering
\includegraphics[width=\linewidth]{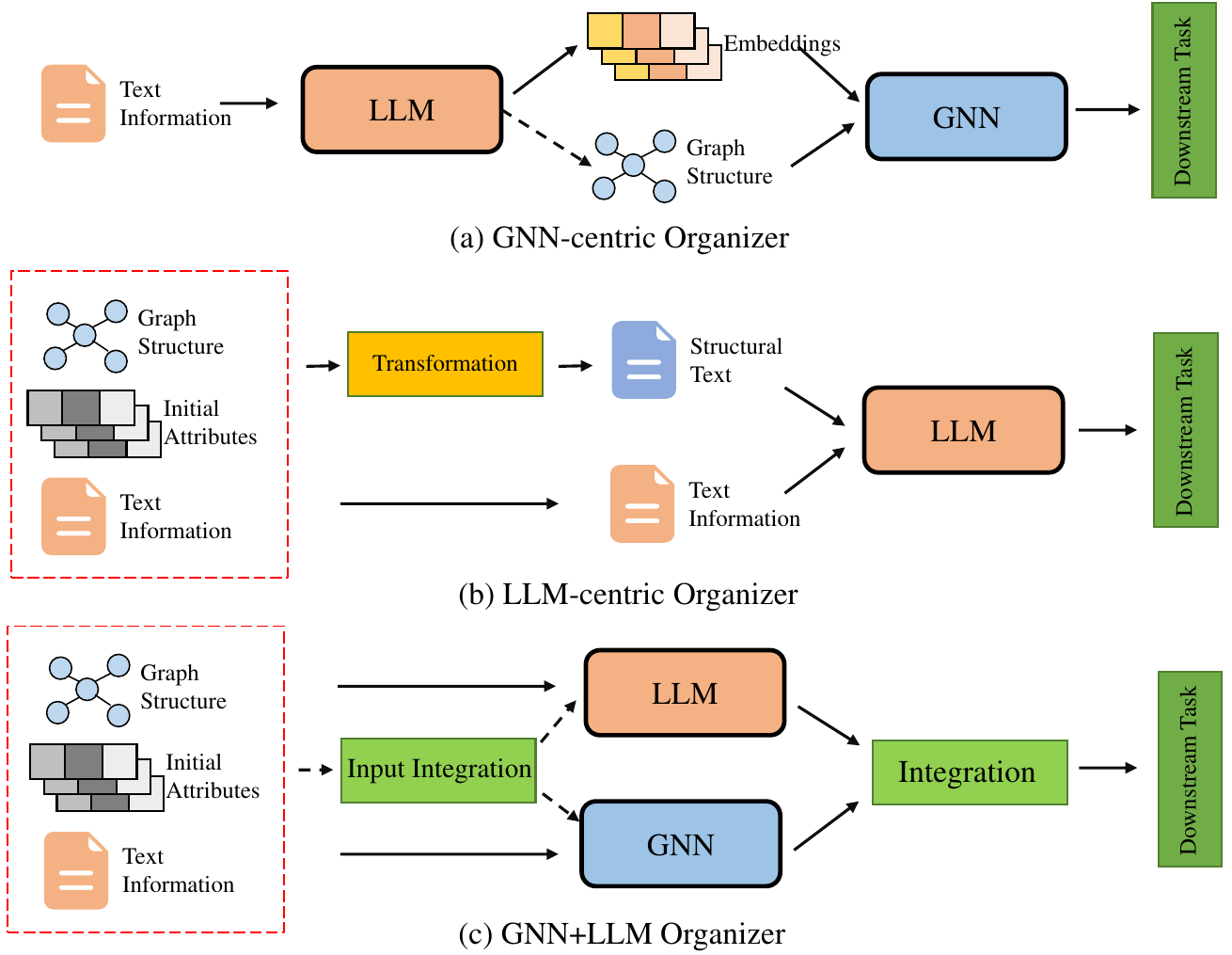}
\caption{The illustration of different knowledge organizers: GNN-centric, LLM-centric, and GNN+LLM organizers.}
\vspace{-3mm}
\label{fig:organizer}
\end{figure}

Beyond the quest for more efficient knowledge extraction, the manner in which extracted knowledge is stored and organized is equally crucial. The advent of LLMs has expanded the design space for knowledge organization systems. Current methodologies combining LLMs with GRL have yielded three distinct types of knowledge organizers based on the integration of GNNs and LLM architectures. In the following, we will delve into an in-depth analysis and discussion of the characteristics and technologies underlying these three types of knowledge organizers: GNN-centric, LLM-centric, and GNN+LLM. The illustration of three knowledge organizers is shown in Figure \ref{fig:organizer}.

\subsection{GNN-centric Organizer}
The focus of GNN-centric Organizer is on using structured encoding ability of GNNs for the final organization and refinement of knowledge. 
In this scenario, LLMs usually act as auxiliary modules to GNNs. They can serve as initializers for node features in GNN inputs \citep{he2023explanations,liu2023one}, optimizers for the graph structures used by GNNs \citep{sun2023large,yu2023empower}, or sources of labels for GNN input data \citep{chen2023label,chen2023exploring}, providing a more comprehensive and sufficient knowledge base for GNN encoding.

In \method{TAPE} \citep{he2023explanations}, LLMs are used to generate additional explanatory and predictive features as inputs for the GNN, with the GNN model ultimately producing the final node representations. In \method{OFA} \citep{liu2023one}, LLMs' versatility is harnessed to encode data from different domains and tasks, enabling GNNs to undergo unified training across various domains and tasks. This breaks the limitation of single GNN models being confined to singular application scenarios, paving the way for the development of large-scale foundational models for graphs.

The improvements and possibilities that LLMs bring to GRL have not been fully explored. Utilizing LLMs as plug-and-play auxiliary modules to enhance GNN models remains a promising and significant direction with a bright future.

\subsection{LLM-centric Organizer}
Recent advancements in GRL have witnessed a growing trend of utilizing LLMs as the core and sole backbone. This paradigm shift is attributed to several key advantages that LLMs offer in seamlessly integrating textual information with graph data and unifying diverse graph learning tasks under a natural language-based framework. Applying LLMs as knowledge organizer to graph modalities involves unique challenges, primarily due to the complexity of transforming graph data into a sequential text format.

\cite{chen2023exploring} developed a method to extract contextual information about a target node using an LLM by constructing the neighbor summary prompt. The output text from this process is then used as a prompt to assist the LLM in tasks related to GRL. 
Structural information is depicted either through the textual features of adjacent nodes or through linguistic descriptions of connectivity relationships. Additionally, the Graph Modeling Language and Graph Markup Language are used to describe graph structure in \method{GPT4Graph} \citep{guo2023gpt4graph}. Furthermore, a method based on tree structures for converting graphs to texts has been proposed in GraphText \citep{zhao2023graphtext}.

Nonetheless, employing LLM-central architecture in GRL presents significant challenges. A primary limitation is the intricacy involved in translating complex graph structures into a format amenable to LLM processing. Moreover, inherent limitations of LLMs, such as their difficulty in managing long-range dependencies and the potential biases embedded within their training data, become pronounced in the context of graph learning. These limitations can affect the model's ability to accurately interpret and utilize the graph-structured information, posing hurdles to effective application in graph-based tasks.

\subsection{GNN+LLM Organizer}
The GNN+LLM Organizer marks a significant advancement in overcoming the limitations of each individual organizer \citep{brannon2023congrat, zhao2022learning, wen2023prompt}. GNN-based models, while adept at structural analysis, fall short in processing textual data and interpreting natural language instructions. Conversely, LLM-based models, despite their linguistic prowess, struggle with precise mathematical calculations and multi-hop logical reasoning. This complementary nature of GNNs and LLMs forms the basis for a hybrid model that leverages the strengths of both: the language understanding capabilities of LLMs and the structural analysis proficiency of GNNs.

How to fully leverage the strengths and compensate for the weaknesses of both modalities, effectively integrating knowledge from these two distinct modalities, remains a key challenge for such methods. This will be analyzed in more detail in the next section, focusing on specific knowledge integration strategies between graph and language domains.

\section{Integration Strategies} \label{sec.integration-strategies}

\begin{figure*}[t]
\centering
\includegraphics[width=\linewidth]{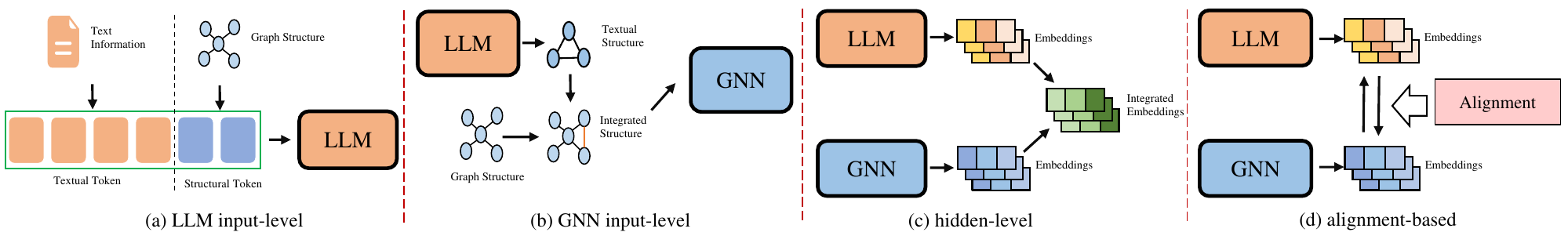}
\caption{The illustration of different knowledge integration strategies.}
\vspace{-3mm}
\label{fig:integrate}
\end{figure*}

Incorporating LLMs into model designs enables the generation of text modality representations for graph data, which effectively complement the structural representations derived from GNNs. For a holistic representation, it is crucial to merge semantic and structural knowledge. Centering on the forms of integration between modal knowledge and representations, we have categorized the current integration strategies into three classes: input-level integration at the input stage, hidden-level integration at the latent processing stage, and indirect integration through alignment, as depicted in Figure \ref{fig:integrate}.

\subsection{Input-level integration}
Input-level integration typically employs a unified model structure to amalgamate graph structure and textual information at the input stage. This integration is formatted to align with the model’s input demands, enabling a seamless integration of data from both modalities. 
Models built upon the Transformer architecture of LLMs typically convert graph structures into natural language narratives, seamlessly blending them with textual data. Conversely, models that pivot on GNNs tend to assimilate textual information by forming virtual nodes that maintain specified structural relations, thereby enabling the GNNs to adeptly manage the synthesis and encoding of structure.

\cite{chen2023exploring} considers first aggregating the textual information of nodes within a neighborhood and then integrating this summarized description into the textual input using a prompt format. \method{InstructGLM} \citep{ye2023natural} integrates diverse hop-level contextual node information with a node sampling strategy into the input of the LLM model. 
Additionally, \method{OFA} \citep{liu2023one}, using GNNs as the knowledge organizer, incorporates textual information into the graph data in the form of prompting virtual nodes. It initializes these virtual nodes with additional task description text, providing not only extra semantic information but also enabling cross-domain multi-task model training integration.

\subsection{Hidden-level Integration}
Hidden-level integration refers to merging the textual semantic representations encoded by LLMs with the graph information representations encoded by GNNs to create a comprehensive representation that fully expresses both semantic and structural information of nodes. 
\method{TAPE} \citep{he2023explanations} employs original textual features, explanatory textual features, and predictive textual features as inputs for GNNs. It uses a straightforward ensemble approach to fuse the predictive representations of three distinct semantic features encoded by three independent GNNs, which capture complementary information from diverse input sources. Additionally, cascading \citep{chandra2020graph} and concatenation \citep{edwards2021text2mol} are commonly used hidden-level integration strategies in previous works to enhance the overall model's ability to capture and integrate diverse aspects of graph data.

Given that current applications of LLM models tend to be quite direct, more sophisticated and effective methods for the integration of textual and graph representations have yet to be explored. How to comprehensively utilize the representations from both modalities to generate higher-order representations with greater expressive capabilities remains an important question to address.

\subsection{Alignment-based Integration}
Beyond merely integrating original input data and hidden layers, another approach, \ie, alignment,  considers the features of GNNs and LLMs as distinct manifestations of the same entity's knowledge in graph and textual modalities, respectively.
By aligning representations across these two modalities, knowledge can be transferred between them, facilitating an indirect form of integration. 
The objective of alignment-based integration lies in maintaining the distinct functionalities of each modality while synchronizing their embedding spaces at a particular stage. The intent is to forge a unified, holistic representation that encapsulates the collective advantages of textual and structural information.

Alignment-based knowledge integration includes three main categories: contrastive alignment, iterative alignment and distillation alignment.
Contrastive alignment \citep{brannon2023congrat} involves treating the graphical representation and textual representation of the same node as positive examples to conduct contrastive learning for knowledge integration. And \method{G2P2} \citep{wen2023prompt} introduces contrastive learning at multiple levels during pre-training, including node-text, text-summary, and node-summary, which reinforces the alignment between textual and graph representations.
For iterative alignment, iterative interaction between the modalities is allowed in the training process for knowledge transferring. For example, \method{GLEM} \citep{zhao2022learning} introduces a novel pseudo-likelihood variational framework to the iterative training process, where the E-step involves optimizing the LLM, and the M-step focuses on training the GNN.
Additionally, \method{GRAD} \citep{mavromatis2023train} implements a distillation alignment approach for aligning dual modalities. Specifically, it employs a GNN as a teacher model to generate soft labels for an LLM, thereby facilitating the transfer of aggregated information.
However, efficiency issues arising from the training of LLMs have limited the application of alignment fusion in real-world scenarios.

\section{Training Strategies} \label{sec.training-strategies}

The knowledge of LLMs is derived from training on massive natural language corpora, yet there exists a gap between this knowledge and that required in the graph learning domain, where capturing structural relationships on graphs is crucial. To bridge this gap, training strategies are designed to enhance LLMs' adaptability to graph data from a model training perspective, which can be categorized into three types: model pre-training, prompt-based training and instruction tuning.

\subsection{Model Pre-training}
The core idea of graph pertaining \citep{hu2019strategies} is to train a model on a substantial dataset to capture general patterns or knowledge, which can then be tailored for specific downstream tasks. In the realm of GRL, it focuses on extracting inherent structural patterns within graph data, paralleling the way language models learn the syntax and semantics of languages. 
Graph Pre-training methodologies are diverse, ranging from contrastive \citep{you2020graph} to predictive/generative \citep{hu2020gpt} approaches, each leveraging the structural and semantic richness of graph data. When incorporating LLMs into GRL, pre-training can also include textual knowledge from language models, and textual pretraining tasks like network-contextualized masked language modeling \citep{jin2023patton} and context graph prediction \citep{zou2023pretraining}. 
Most current methods integrating LLMs treat them as plug-and-play modules, with relatively few studies focusing on injecting graph structure knowledge into LLMs during pre-training. \method{InstructGLM} \citep{ye2023natural} pioneered the use of self-supervised link prediction task as the auxiliary training task for LLMs to comprehend graph structural knowledge. The significant performance indicates that considering a pre-training paradigm that blends GNNs and LLMs integration is both necessary and feasible.

\subsection{Prompt-based Training}
Prompt-based training comprises Prompting and Prompt Tuning \citep{liu2023pre}. The former guides language models to produce specific outputs, while the latter focuses on aligning downstream tasks with pre-training tasks. The introduction of LLMs has led to the widespread use of prompting to enhance models' understanding of graph structural information. Neighborhood or connection information is added as prompting to improve LLMs' adaptability to GRL \citep{yu2023empower, chen2023label}, or to activate their few/zero-shot capabilities \citep{guo2023gpt4graph, huang2023can}.

For graph tasks, the concept of graph prompts has been explored extensively, aiming at the integration of diverse graph tasks. 
For example, \method{GPPT} \citep{sun2022gppt} reconceptualizes graph tasks as edge prediction problems, while \method{GraphPrompt} \citep{liu2023graphprompt} extends this framework by unifying tasks as subgraph similarity calculations. While \method{OFA} \citep{liu2023one} employs both prompting and graph prompt tuning. Prompting are used for feature dimension alignment and initializations of nodes across different datasets, and graph prompt tuning is used to unify different tasks, enabling a single model to be trained and evaluated across various datasets and tasks. 
In the field of GRL with LLMs, where large-scale data is often scarce, prompt-based training remains a vital technique. 

\subsection{Instruction Tuning}
The core methodology of Instruction Tuning involves integrating the pre-trained models' input data with task-specific instructional prompts. Instruction Tuning is executed within a multi-prompt training framework, where the model is exposed to various task instructions, aiding in its understanding and response to diverse task requirements. 

In the realm of GRL, where annotated data is relatively limited and downstream tasks often span multiple domains and objectives, Instruction Tuning is particularly valuable for enhancing model performance in few-shot and zero-shot scenarios. This efficient fine-tuning approach can effectively tap into the inherent knowledge related to GRL within LLMs, thereby enhancing their comprehension abilities for graph-related tasks. 
\method{InstructGLM} \citep{ye2023natural} directs the LLM to generate responses for various graph learning tasks within a unified language modeling framework. 
\method{GraphGPT} \citep{tang2023graphgpt} proposes a two-stage instruction tuning approach for graph learning. Initially, it uses self-supervised tasks to teach the LLM graph structure knowledge. Then, it applies task-specific tuning to improve the LLM's performance on downstream tasks, aligning it with graph domain knowledge and specific task requirements.
As research on GRL with LLMs progresses, instruction tuning will play an increasingly crucial role in fine-tuning LLMs.


\section{Future Directions} \label{sec.future-directions}
In this section, we explore potential avenues for future research of advancing GRL \wrt our organization framework.

\stitle{Generalization of knowledge extractor.}
Current progress indicates using LLMs for text-attributed graphs has been promising, but challenges arise with graph data lacking rich text. Adapting LLMs to interpret non-textual graph data is crucial for progress in GRL, especially considering the ubiquity of non-textual graphs in real-world scenarios.

\stitle{Effectiveness of knowledge organizer.}
The development of LLMs has led to the coexistence of three distinct architectures in the realm of Graph Foundation Models: GNNs, Graph Transformers, and LLMs. However, there is no consensus on how to design an effective knowledge organizer for GRL, and a unified theoretical framework to analyze the strengths and weaknesses of these various architectures is lacking. 

\stitle{Transferability of integration strategies.}
Transferability in graph learning is challenging due to the unique characteristics of each graph, such as size, connectivity, and topology, which makes it difficult to apply knowledge across different graphs.
However, integrating with the exceptional generalization capabilities of LLMs offers potential solutions to the challenges of transferability.
Advancing transferability requires not just subtle integration strategies but also a deeper understanding of knowledge transfer in graph domain. 

\stitle{Adaptability of training strategies.}
Furthermore, there is a scarcity of research on the pre-training and fine-tuning techniques of LLMs on graph data. Effective training strategies need to consider LLMs' methods and structures and how to integrate graph information, posing a significant challenge in adapting LLMs to graph learning efficiently.

\section{Conclusions} \label{sec.conclusions}
In this survey, we systematically reviewed and summarized the technical advancements in GRL with LLMs.
We provided a comprehensive analysis of the essential components and operations of these models from a technical perspective.
We then proposed a novel taxonomy that categorizes literature into two main components: knowledge extractors and organizers and covers operation techniques of integration and training strategies.
We further identified and discussed potential future research directions to suggest avenues for further development.

\newpage
\bibliographystyle{named}
\bibliography{main}

\end{document}